RESEARCH

# Meta-ethics and AI: exploring the novel meta-ethical questions in the era of AI

**Shang Lu**[1]

**Abstract**

With the development of artificial intelligence (AI), the landscape of meta-ethics, which has largely centred on human ethics, faces pressures that may significantly reconfigure it. In particular, if future AI systems were to exhibit sufficiently integrated capacities for moral reasoning, moral intentionality, and moral reflection, novel meta-ethical questions would arise concerning what I call 'AI's own ethics,' as distinct from ethical principles merely imposed on AI by human designers. This paper offers a conditional and methodological framework for identifying the questions that would emerge if such AI systems were to arise. On that basis, the paper distinguishes four domains of meta-ethical inquiry in the era of AI: questions about the nature of human ethics from the human perspective; questions about the nature of AI's own ethics from the human perspective; questions about the nature of human ethics from the AI perspective; and questions about the nature of AI's own ethics from the AI perspective. The paper then considers how some existing mainstream meta-ethical theories—such as cognitivism and non-cognitivism, error theory and success theory, relativism, and objective realism—might illuminate these domains, while arguing that many familiar human-centred formulations of those theories may not transfer straightforwardly to AI cases without substantial revision. The overall conclusion is that the emergence of AI's own ethics would place significant pressure on current frameworks and may require substantial refinement, reconstruction, or reconceptualisation.

## 1 Introduction

We are, if not already deeply in, at the beginning of the era of AI. On the one hand, AIs are already making significant decisions (such as autonomous driving systems and combat AI drones) and participating in moral discourse (such as large language models), even though they may not yet possess human-level intelligence. On the other hand, the possibility of artificial general intelligence (AGI), alleged to possess human-level intelligence, and even artificial superintelligence (ASI), alleged to surpass human intelligence, is no longer merely science fiction. According to recent predictions from more optimistic experts, although some believe that AGI is not imminent [61], the first emergence of AGI could occur as early as the 2040s [1, 79], depending on how it is defined. Even the more conservative experts estimate that AGI will emerge between 2050 and 2075 [49]. Moreover, even though large language models might not be the path toward achieving AGI, there could be other viable pathways, such as the world models [71] and embodied intelligence [45, 123].

As AI drastically changes people's everyday lives, it deeply shifts conventional ethical paradigms. For example, historically, moral agency has been ascribed exclusively to human beings—beings capable of intentionality, deliberation, and moral responsibility. With AI systems increasingly participating in decision-making processes—ranging from autonomous vehicles to healthcare diagnostics—the locus of moral responsibility is being reconsidered in the literature [122]. Similarly, the integration of AI into everyday life compels us to reconsider the ethical implications of our interactions with non-human agents. This shift is not merely about attributing moral responsibility to machines but also involves rethinking how human values are maintained or transformed in an increasingly automated world [51]. As AI systems assume roles traditionally reserved for human judgment (e.g., moral advisors, high-stakes decision-makers), debates over the accountability, transparency, explainability, and trustworthiness of non-human agents are becoming central to contemporary moral discourse.

✉ Shang Lu
shang.lu@unsw.edu.au

[1] School of Computer Science and Engineering, UNSW Sydney, Sydney, Australia

By contrast, the discussions related to AI in meta-ethics—the reflective meta-discourse of ethics—are so far relatively underdeveloped (although there are a few discussions on AI-related applied meta-ethics, which I will get into in the next section). The reason, I suspect, is that although the priority of applied and normative ethics may have shifted due to the emergence of AI, most ethical questions surrounding AI, as mentioned above, remain within the *scope* of conventional moral discourse. Although these questions are often claimed to be about 'AI ethics' or 'machine ethics,' they still belong to the ethics of certain groups of humans, or what one may call the 'AI designers' ethics' or 'machine engineers' ethics.' Hence, the subject matter of meta-ethics, i.e., the nature of ethics as a discourse, has not undergone significant change.

However, the landscape of meta-ethics may shift dramatically as AI continues to develop. This is not merely because new normative questions related to AI will arise, but also because there might be ethical systems that differ from human ethics, offering additional perspectives for examining those different kinds of ethics. The canonical focus of meta-ethics has been on describing the nature of human moral discourse and analysing its profound differences, if any, from other areas, such as the natural sciences. With the development of AI, many new meta-ethical questions would emerge, given the possibility of 'AI's own ethics'—the ethical faculty of AIs themselves, rather than the mere ethical principles imposed or programmed into AI by human engineers.

Those questions include, for example: What does it mean for AIs to have their own ethics? What would be the nature of AI's own ethics? Should we, and to what extent, implement ethical and meta-ethical theories to initiate AI's own ethics? What is the nature of human ethics to AI with their own ethics? Whether—and if so, how—is it different from the nature of human ethics to ourselves?

This paper aims to identify and categorise those novel meta-ethical questions in the era of AI. I argue that there will be four main categories of meta-ethical questions in the era of AI, and three of them will be closely related to both AI's own ethics and AI's meta-ethical perspective. I also examine how existing mainstream meta-ethical theories, including non-cognitivism, error theory, relativism, and objective realism, may illuminate these domains, while arguing that many familiar human-centred formulations of those theories may not transfer straightforwardly to AI cases without significant refinement, reconstruction, or reconceptualisation.

The structure of the paper is as follows. In Sect. 2, I distinguish the so-called 'AI ethics' in the current literature from AI's own ethics and introduce a working threshold for attributing the latter. In Sect. 3, I organise the relevant meta-ethical questions into four domains. In Sects. 4–7, I explore representative questions in each domain and consider how existing mainstream meta-ethical theories might apply to them, as well as where their current formulations may encounter difficulty. Section 8 concludes by drawing out the broader implications for meta-ethics in the era of AI.

## 2 The difference between AI ethics and AI's own ethics

### 2.1 Three conditions for attributing AI's own ethics

I begin by distinguishing the so-called 'AI ethics' in the current literature from what I will call 'AI's own ethics.' The term 'AI ethics' usually refers to the ethical guidelines, constraints, and decision-making frameworks built into AI systems—what is often discussed under the heading of machine ethics. Machine ethics aims to design AI that can autonomously avoid harmful actions and, ideally, operate in ways consistent with normative ethical principles and other human values. For instance, efforts in machine ethics typically aim to embed constraints in AI algorithms so that its outputs remain aligned with human values [51, 122].

However, such 'AI ethics' is ultimately a reflection of the moral principles of its human creators rather than a genuine morality on the part of the machine itself. In other words, the current so-called AI ethics is merely an extension of the ethics of AI designers (and sometimes AI users as well). It is thus completely dependent on the moral principles and constraints the engineers implement to the algorithm during processes such as reinforcement learning from human feedback (RLFH), whereby a reward model learned from human raters guides the policy toward responses deemed preferable by annotators [24, 87], pretraining corpus, and 'value alignment' research [43]. AI ethics is more complex than other machine ethics, as AI designers often cannot predict exactly what the AI they design will do in many specific situations. However, fundamentally, it is still the engineers' decision what ethical framework and values their products abide by and how to train them to align with those values. Therefore, the current ethical investigation in AI ethics does not appear to generate many original meta-ethical questions.

By contrast, the expression 'AI's own ethics' is meant to pick out a different possibility: not merely ethics imposed on AI from outside, but a case in which an AI system exhibits an internally organised ethical standpoint of its own. I do not claim here that our current ordinary concept of morality already entails that such a case is possible, nor that a purely functional criterion settles the deeper metaphysics of moral subjecthood. Rather, the proposal in this section is methodological and conditional. It offers a working explication of 'AI's own ethics' for the limited purpose of identifying the novel meta-ethical questions that would arise if future AI systems were to exhibit sufficiently integrated moral capacities. On this approach, the relevant issue is not whether AI already fits unmodified human concepts of moral agency, but what threshold would make the attribution of an ethical standpoint theoretically warranted within this inquiry. More specifically, I propose to treat an AI system as possessing AI's own ethics, for the purposes of this paper, when it exhibits three functionally integrated capacities: moral reasoning, moral intentionality, and moral reflection.

First, the AI must be capable of moral reasoning—the capacity not only to infer what is morally right or wrong based on provided normative frameworks, but also to derive the moral principles and theories it should abide by. The capacity for merely generating moral judgments (however 'satisfying' they are) does not count as moral reasoning. For example, although many current large language models (LLMs) can generate texts that appear to be moral judgments, and many people do believe they can sometimes serve as efficient moral advisors for everyday moral inquiries, they are not capable of moral reasoning. That is because, while making apparent 'moral judgments' by efficiently predicting the next token in a moral sentence, the current LLMs are not capable of deriving what moral principles they should abide by themselves. As a result, the processes of current LLMs generating moral judgments should not be seen as processes of moral reasoning.

Second, the AI must possess moral intentionality—intending to do what is morally right and to avoid what is morally wrong—grounded in its own moral principles and theories derived from its own reasoning. The capacity to merely follow specific moral rules and principles, as implemented in processes such as the 'top-down' training—where the designer deliberately establishes a goal for the machine from the beginning, grounded in the moral principles they intend to operationalise [122]—does not count as moral intentionality. For example, the fact that automated driving vehicles tend to avoid running over anything that appears to be a human is not evidence that these vehicles have moral intentionality, as the essential reason for these behaviours is preprogrammed rules. Similarly, current LLMs do not demonstrate moral intentionality solely because they refrain from assisting with harmful objectives under externally imposed constraints.

Third, the AI must have moral reflection—the capacity for critically reflecting on its own moral frameworks, allowing for revision when inconsistencies are detected [93]. The capacity for merely adjusting its outputs or decision pattern based on external feedback received in processes such as the 'bottom-up' training—where the designer creates an environment or feedback systems for AI to 'learn' from human behaviours by being 'rewarded' for morally praiseworthy conducts and being 'punished' for blameworthy ones [122]—does not count as moral reflection. For example, the fact that AI for automated military drones gets better at avoiding killing field medics and reporters during simulated RLFH combat training is not a sign that the AI is capable of moral reflection. That is because what the AI is reinforced during those trainings is merely a mechanism for earning more 'reward' points and fewer 'punishment' points, not a revision of its own moral frameworks.

As used here, these conditions are specified functionally rather than phenomenologically. Accordingly, the framework does not build human-like consciousness, beliefs, desires, or feelings into the threshold criterion for attributing AI's own ethics,[1] while leaving open whether such features are required for stronger notions such as full moral subjecthood, responsibility, or standing. It is offered as a threshold criterion for distinguishing systems whose apparent moral outputs merely reflect externally imposed rules or training from systems whose moral activity is better explained by internally organised morality.

This criterion also aligns with the focus on capacities rather than processes in defining significant computing research terms, such as 'Artificial General Intelligence (AGI)' and 'Artificial Superintelligence (ASI).' According to these definitions, it is essential to focus on the tasks an AI can perform rather than the underlying processes by which it performs them [81]. That is to say, an AGI system is capable of performing a task as long as it possesses the relevant functional capacity, even without thinking or understanding in a human-like way or possessing traits such as consciousness [20] and sentience.

These three capacities are necessary because, without moral reasoning, an agent's moral behaviour is simply rule execution or pattern matching. Intentionality supplies the directedness that makes behaviour morally attributable to the agent rather than to externally imposed directives. And without reflective capacity, an apparent pattern of moral behaviour cannot be said to be owned by the agent, since ownership requires the capacity for self-attribution and revision. Empirically, agents that lack any one of these capacities are routinely treated by scientists, ethicists, and lay observers as morally guided tools rather than moral selves. Conversely, when all three are present and empirically indispensable to explaining the agent's behaviour, denying that the agent possesses its own morality appears ad hoc.

[1] In other words, what matters is whether the relevant agential capacities are instantiated, not whether they are instantiated in a human way. See [6].

Below, I present a brief, empirically grounded analogy to support the claim that moral reasoning, intentionality, and reflective capacities together also sufficiently constitute the possibility that an agent possesses its own morality. The analogy proceeds through a practical shift in how we attribute 'insight and understanding' to artificial players in complex board games.[2]

When Deep Blue defeated Garry Kasparov in 1997, the immediate and widespread judgment was not that the machine genuinely knew how to play chess in the human sense, let alone that it had any insight and understanding of the game. Observers explained Deep Blue's success by pointing to its brute-force search, its reliance on human-designed evaluation functions, and the lack of adaptive, internally generated strategies. Empirically, Deep Blue's behaviour was best modelled as the application of externally supplied rules and massive computation over a well-specified state space; its impressive outcomes did not warrant attributing a self-governing chess competence.

By contrast, when AlphaGo and its successors began to defeat top human players in Go—first with human-guided learning [103] and later via self-play that produced novel strategies [104]—the pattern of explanation shifted. AlphaGo's play exhibited genuine novelty, internal learning processes, and strategies that reshaped human practice: professional players began studying and adopting moves and patterns first discovered by the system. In this case, it seems difficult to deny that what AlphaGo 'knows' is neither mere rule knowledge nor abstract theory, but a structured form of practical rationality—exactly the sense in which AlphaGo is often said to have grasped the 'insight and understanding of Go' more deeply than human players. The best empirical explanation for this shift is not merely a larger or more opaque lookup table but the presence of a flexible, generalisable ability that manifests in behaviour, learning, and influence on human experts.

This empirical pattern in the domain of games generalises in an illuminating way to AI's ethics. We can, first, distinguish between agents whose moral behaviour is derivative and rule-based (the counterpart of Deep Blue) and agents whose moral behaviour requires positing autonomous, internally structured capacities (the counterpart of AlphaGo). An agent that reliably outputs correct moral judgments or behaves in morally appropriate ways by implementing fixed rules, reward-maximising heuristics, or externally programmed mappings remains explanatorily analogous to a chess engine that uses precomputed lines. In such cases, our empirical warrant supports attributing moral *conformity*—the agent acts in accord with norms—but not moral *possession* or ownership. By contrast, when an agent's moral responses instantiate novelty across novel cases, when the agent develops principles through processes not reducible to externally supplied prescriptions, and when its moral behaviour causally influences and is learnable by human moral agents, the best empirical explanation invokes internal capacities for moral reasoning, intentional representation, and reflective endorsement. At that point, attributing the agent's own morality becomes the more parsimonious and explanatorily robust option.

Just like AlphaGo does not conceptually need to 'learn' the insight and understanding of Go playing in the human sense to possess it (AlphaGo makes a move because it calculates that the move is the one that will increase winning probability to the maximum, not because the move abides by any pre-established principles of strategies), it is also not conceptually determined that a moral AI must possess any consciousness or mental states in the human sense to function its own moral capacities and thus possess its own morality. In fact, ordinary moral attribution in human social practice does not hinge on direct access to another's phenomenology either; it hinges on behaviour, reasons-responsiveness, and the capacity for justificatory dialogue and revision. Empirically, then, phenomenal consciousness is not a necessary empirical marker for attributing moral ownership in the social sense—although one may argue it is metaphysically or normatively relevant for deeper forms of moral responsibility.

It is, however, significant to note that the analogy is not presented as a conclusive conceptual proof that AI can possess morality without consciousness or other mental states. Nor are they meant to establish that explanatory attribution is identical with metaphysical parity. It is intended only to support the narrower methodological point that, where externally imposed rule-following no longer provides the best explanation of a system's moral behaviour, a functional attribution of AI's own ethics may become theoretically warranted. Whether stronger notions such as full moral agency, moral responsibility, or moral standing require consciousness, phenomenology, embodiment, or other psychological features remains a further question, and one that this paper does not attempt to settle.

The conditions for AI's own ethics may still encounter resistance from both ends. An engineer may still worry about their necessity due to potential 'anthropomorphism', while a philosopher may still worry about their sufficiency due to thought experiments such as the 'philosophical zombie.' In what follows, I will briefly address the relevant objections from both ends.

[2] The best term for the 'insight and understanding' of playing board-games, such as chess, Chinese chess and Go, is the Chinese word '棋理', which does not admit of a single perfect English equivalent, because it occupies an intermediate conceptual space between principles, insight, both theoretical and practical understanding of a board game. In the following, I will simply use the term 'insight and understanding (of a board game)' for '棋理'.

The first objection concerns whether attributing 'its own morality' to AI based on observed moral reasoning, intentionality, and reflection risks anthropomorphism, i.e., projecting distinctly human mental qualities onto artificial systems. Opponents may suggest that because current AI lacks consciousness, subjective experience, or biological embodiment, attributing genuine moral cognition to it conflates surface functional analogy with ontological parity [101].

However, this worry rests on the assumption that the explanatory ascription of AI's own morality entails the metaphysical attribution of human-like phenomenology. The functionalist stipulation in the definition of AI's own ethics explicitly denies this: it defines capacities in terms of system-level functional roles (deriving principles, goal-directed normativity, meta-level revision), rather than in terms of subjective phenomenology. This distinction matters because functional descriptions are multiply realisable: a capacity can be instantiated by human neural processes or by mechanistic algorithms without implying common inner qualia [29]. Thus, to claim that an AI satisfies the three conditions is to claim that it bears systems-level relations and capacities, not that it shares human consciousness.

More specifically, first, an AI can perform moral reasoning through functions such as rule formation, consistency checks, and principled generalisation without invoking subjective experience; the claim is methodological: we treat these organised operations as genuine instances of reasoning when they meet the functional criteria [122]. Second, Bratman's account of intention as plan-structured practical reasoning shows how intention can be analysed without immediate appeal to phenomenology: intentions organise action and guide deliberation [18]. Interpreting AI intentionality thus involves attributing structured, normative action-guiding states, which is an explanatory stance about system organisation rather than a claim that the system has human-like desires or feelings [29]. Third, moral reflection, as a revisionary capacity, resembles formal mechanisms of meta-level evaluation and belief revision (e.g., adopting Rawls's reflective equilibrium in method). When an AI detects inconsistency among its normative outputs and systematically restructures its decision procedures or principles, we have functional reflection: a meta-procedural correction that need not presuppose subjective reflection [93, 122]. Finally, recent work in machine ethics emphasises explicit distinctions between artefactual moral competence and moral personhood, recognising functional moral capacities in artefacts is compatible with denying moral personhood or inner consciousness [37].

In short, insofar as the three conditions are read as functional, mechanistic criteria for explanatory ascription, they do not entail anthropomorphism; they instead permit a careful scientific vocabulary for distinguishing levels of moral competence without metaphysical overreach.

The second objection asks whether sophisticated mimicry (e.g., a philosophical zombie[3]) might impersonate all three capacities. The response is to insist on explanatory parsimony: if the simplest explanation of an agent's sustained, novel, and causally efficacious moral behaviour is the possession of internal capacities for reasoning, intentionality, and reflection, then we are warranted in attributing moral ownership even if, in principle, a perfectly engineered mimic could produce the same outward signs. Scientific practice routinely treats such inferences as legitimate: attribution rests on whether internal capacities are the best explanation of observed behaviour, not on the impossible demand to rule out any hypothetical mimic.

Philosophical zombies further show, at least conceptually, that full behavioural, reasoning, intentional, and reflective capacities could in principle exist without subjective experience; if that is so, then our empirical criteria—moral reasoning, intentionality, and reflection—do not logically presume phenomenal consciousness. From an empirical-attribution standpoint, this supports the move I defended earlier: moral ownership can be warranted by the best explanation of behaviour and internal functional capacities even when we cannot independently access phenomenology. However, opponents may still reply that moral ownership (or full moral status and responsibility) *normatively* depends on subjective experience and that a being functionally identical to us but lacking experience would not truly count as a moral subject. That rejoinder is substantive and cannot be settled by empirical observation alone; it converts the debate into a normative and metaphysical dispute about whether phenomenal consciousness is part of what it is to be a 'complete' moral agent rather than merely an explanatorily useful correlate of moral agency. It thus needs to be addressed separately and is beyond the scope of this paper.

To summarise, this section has distinguished 'AI's own ethics' from ethical principles merely imposed on AI by human designers and institutions. Rather than offering a final metaphysical analysis of moral subjecthood, it has proposed a working threshold for attributing AI's own ethics within the purposes of this paper. On this approach, an AI system may be treated as possessing its own ethics when its moral activity is better explained by the integrated presence of moral reasoning, moral intentionality, and moral reflection than by externally imposed rules or training alone. This proposal is functional in a limited sense: it identifies a threshold for attribution by reference to the roles these capacities play, rather than by reference to a particular biological substrate or to explicitly human forms of consciousness, desire, or feeling.

[3] A 'philosophical zombie' is a hypothetical being physically and functionally identical to a normal human but lacking phenomenal consciousness—that is, there is nothing it is like for the zombie to be that subject. In thought experiments originating in the contemporary discussion of consciousness, most prominently deployed by David Chalmers, zombies behave, report, and function exactly as ordinary persons do, yet they have no subjective qualitative experience (see [21]).

## 2.2 The meta-ethical neutrality of the conditions

A further clarification is now needed. The three conditions proposed above are not fully neutral in every theoretical respect. They presuppose a threshold criterion for attributing AI's own ethics, and that criterion is functional in the limited sense just noted. However, this limited commitment should not be mistaken for a full resolution of the downstream meta-ethical questions. The claim of the previous section was only that, for the purposes of this paper, moral reasoning, moral intentionality, and moral reflection provide a useful threshold for distinguishing externally imposed moral conformity from a more internally organised ethical standpoint. It was not that this threshold already determines the semantics, ontology, epistemology, or psychology of the resulting ethical discourse.

Accordingly, the functional conditions should not be read as against non-cognitivism, error theory, or other anti-realist positions. The point is not that those views are excluded by definition. Rather, the question is whether familiar human-centred formulations of the major meta-ethical theories can adequately characterise a case in which AI is treated as having its own ethics under the threshold criterion proposed here. Put differently, the first question is when it is theoretically warranted to attribute AI's own ethics at all; the second is how such ethics, once attributed, should be meta-ethically understood. The first question does not settle the second.

With that distinction in place, one can now ask in a more disciplined way whether demanding moral reasoning presupposes cognitivism, whether demanding moral intentionality presupposes internalism, whether demanding moral reflection presupposes success theory, and whether the possibility of AI's own ethics entails relativism. My claim is not that the conditions are wholly theory-free, but that they do not by themselves force us to resolve those further disputes in advance.

Firstly, demanding the capacity for moral reasoning for AI's own ethics does not presume meta-ethical cognitivism. Meta-ethical cognitivism is the view that moral statements, just like statements in most other mainstream discourses, such as physics and medicine, are truth-apt. Meta-ethical non-cognitivism, by contrast, is the theory that moral statements are not in the business of being either true or false.

It may appear that moral reasoning conditions on the presumptive fact that moral judgments are truth-apt. However, non-cognitivism does not prohibit moral reasoning. For example, one could argue that moral reasoning exhibits a logic of attitudes, according to which logical connectives such as conditionals express higher-order attitudes towards accepting certain conjunctions of attitudes [10]. One could also adopt a variety of non-cognitivism (e.g., quasi-realism,see [11]), according to which moral statements can be either true or false in a minimal sense but not truth-apt in a more robust, non-minimal sense, thereby allowing the possibility of moral logic itself [111]. Hence, demanding that AI possess the capacity for moral reasoning in its own ethics does not presuppose a stance between cognitivism and non-cognitivism.

Secondly, demanding the capacity for moral intentionality does not presume motivational internalism. Motivational internalism is the meta-ethical view that there is a *necessary* (conceptual or metaphysical) link between a sincere moral judgment about an action (e.g., 'stealing is wrong') and at least a minimal motivation (not) to perform that action (stealing) [108]. Its opposing theory, motivational externalism, denies any *necessary* connection between moral judgment and motivation. One can fully comprehend and sincerely endorse a moral judgment without any associated motivation [19]. On this view, moral judgments are only contingently connected with motivation.

Moral intentionality is a distinct concept from moral motivation. A moral intention is a plan-like propositional state whose content is explicitly moral: it represents an agent as committed to a future course of action for a moral reason. Unlike mere desires, intentions incorporate a temporal structure (planning, sequencing) and a normative commitment to carry out what one has resolved to do [18]. Moral motivation denotes the set of desires, drives, or normative impulses that provide the 'push' to perform a moral action [39, 82]. One can intend morally without presently feeling strong moral motivation (e.g., intending to keep a promise despite low desire owing to a sense of duty) [3]. One can also feel morally motivated without forming a concrete moral intention (e.g., feeling compassion for the homeless without forming a specific plan to help) [126]. Similarly, an AI can intend morally without any feelings or desires associated with moral motivation. One could be a motivational externalist while still believing that an AI can commit to a future course of action based on its own moral judgments, even though those judgments do not, in and of themselves, motivate it. Therefore, demanding moral intentionality for AI's own ethics does not presume a stance between motivational internalism and externalism.

Thirdly, demanding the capacity for moral reflection in AI's own ethics does not presuppose a success theory of moral statements. Success theory of moral statements is the opposing theory to moral error theory. Moral error theory posits that moral statements are systematically false, often supported by the argument that moral discourse presumes the existence of objective moral properties and facts, whereas there is no such thing [77]. Success theory, in opposition to error theory, states that *some* moral statements are true, either because moral discourse does not presume objective moral properties or facts, or because there are objective moral properties and facts.

Moral reflection does not presume either success or error theory. Moral reflection typically involves critical evaluation of one's received moral convictions against reasons and counterexamples [82]. Under success theory, moral reflection is a form of truth-seeking in which an agent examines whether their moral judgments track objective moral facts [19], and error correction is achieved by testing moral hypotheses against evidence of harm, rights, and other relevant considerations [69, 99]. By contrast, moral reflection under error theory can take the form of analysing the structure and function of moral statements (their expressive, prescriptive, or projective roles) [12, 46] and assessing moral practices in terms of their consequences, such as social effects (e.g., promoting cooperation, stabilising society), rather than their truth-conditions. Therefore, demanding moral reflection for AI's own ethics does not presume a stance between moral error theory and success theory.

Fourthly, the assumption that AI's own ethics, with a very different nature from human ethics, could emerge as its capacities for moral reasoning, intentionality, and reflection develop does not entail moral relativism. Moral Relativism holds that the truth value of moral judgments varies for agents with different cultural, social, or individual perspectives; most importantly, there are no universal moral truths [59]. Its opposing view, moral universalism (sometimes vaguely referred to as 'moral objectivism'), maintains that some moral claims are universally true—true for everyone, regardless of cultural or social background [19].

It seems that AI may develop peculiar moral frameworks, as the nature of AI's own ethics would differ from humans' due to their highly divergent origins and processes in moral reasoning, intentionality, and reflection. It may appear that accepting those processes as moral processes and accepting AI-derived judgments from them as moral judgments already commits one to a kind of moral relativism. However, recognising distinct natures of ethics does not necessarily lead to a relativistic stance on the truth value of moral judgments. On the one hand, for the sake of argument, even if humans and AI converge on the truth value of few moral statements, it does not contradict universalism, because either side could be very mistaken in their moral judgments. On the other hand, it is also possible that human ethics and AI's own ethics may converge on some moral judgments because of a significant shared feature—the capacity for moral reflection. Moral reflection involves testing one's moral judgments against reasons, counterexamples, and coherence constraints. Such standards can be articulated universally [83, 99], rather than relative to each entity. Given such standards, human agents with diverse starting points, such as different cultural backgrounds, can converge on similar ethical conclusions when given access to the same evidence and reasoning processes, as demonstrated by empirical work in moral psychology [13, 54]. If moral universalism is true, an AI with a developed capacity for open reflection could likewise converge on some universal moral judgments, at least before the periods when ASIs significantly surpass humans in moral capacities.

Therefore, the proposed definition of AI's own ethics as requiring moral reasoning, moral intentionality, and moral reflection should not be read as already settling the subsequent meta-ethical questions. It is intended only as a criterion for the possession of AI's own ethics at all. This criterion is functional in the limited sense that it concerns which capacities a system must instantiate to count as having its own ethics, rather than what moral judgments ultimately are, whether they are truth-apt, or whether moral facts exist.

Another point worth mentioning is that the definition of AI's own ethics also does not commit one to any *normative* theory about AI's moral status. To be more specific, the idea that an AI develops its own ethics does not, in itself, imply that the AI possesses moral standing, responsibility, rights, or moral agency; whether it has these features depends entirely on the relevant normative theory. For example, if moral standing merely requires moral rationality, then an AI with the capacity for moral reasoning, intentionality and reflection certainly has a moral standing. By contrast, if the true normative theory holds that only sentient beings—beings that can suffer and experience pleasure—have moral standing [8, 106], then an AI does not possess moral standing merely because it has an ethical dimension. Similarly, if respect for others' rights is a sufficient condition for having rights, then an AI with its own ethics that recognises rights (possibly human rights) should also have rights. By contrast, if to have a right to something means that one must at least be capable of desiring it [117], then it is not guaranteed that an AI with its own ethics has rights, since having a moral faculty for AI does not necessarily link to having human-like psychological states such as desires. Moral agency typically conditions on moral responsibility, and moral responsibility, in turn, conditions on moral autonomy. Again, if a certain level of functional capacities, such as moral representation, reflective deliberation, and openness to revision and critique, is sufficient for autonomy, then AIs with their own ethics may be considered moral agents. However, if autonomy also requires a psychological sense of spontaneity or 'freedom,' then probably no AI can be deemed a moral agent.

To summarise, the three conditions are not wholly neutral, since they presuppose a functional threshold for attributing AI's own ethics. However, this limited commitment does not settle the later meta-ethical questions. Its role is only to mark, for the purposes of this paper, the point at which AI may be treated as exhibiting its own ethical standpoint rather than mere externally imposed moral conformity. Once such a case is in view, further questions remain open concerning the semantics, ontology, epistemology, and psychology of that ethics.

## 3 The four domains of meta-ethical questions in the era of AI

Since current AI systems are only capable of following some 'top-down' human-imposed ethical guidelines or acting apparently aligned with specific normative values (to a degree) as a result of 'bottom-up' RLFH training, they do not possess their own ethics. However, as AI's capacities become comparable or even superior to those of humans in various areas, there seems to be no obviously non-ad hoc reason not to presume that this will also be the case in ethical capacities.

In this paper, I make the empirical assumption that the emergence of AI's own ethics is at least possible in the near future. The possibility of AI's own ethics emerging will raise novel meta-ethical questions in several domains. That is because, firstly, AI's own ethics may have a distinct origin from human ethics. There are, of course, various theories about the origin of human ethics. For example, evolutionary theory suggests that human morality evolved through natural selection. Morally associated human emotions, such as empathy and reciprocity, are biological adaptations that have evolved over time. Kin selection [55] and reciprocal altruism [118] explain cooperation as fitness-maximising strategies for both individual and group survival. By contrast, empiricist theory argues that human morality is constructed through socialisation, cultural norms, and individual learning. Biological predispositions (e.g., empathy) are raw materials shaped by environmental inputs, such as emotionally charged cultural practices, rather than innate modules [90]. Although debates remain about whether human morality is a direct adaptation (e.g., for cooperation) or a byproduct of other traits (e.g., intelligence), and whether cultural evolution can transcend biological constraints (e.g., expanding moral circles to include animals or AIs), most contemporary theorists may propose a 'dual-inheritance' framework about how biology and culture coevolve to shape human moral systems [17, 26].

However, it seems difficult to directly adopt any of these theories about the origin of human ethics to explain the emergence of AI's own ethics. As mentioned in the previous section, AI's own ethics can be broken down into the functional capacities that an AI possesses in terms of moral reasoning, intentionality, and reflection. Those capacities are highly programmed without the necessary links to morally associated emotions or environmental inputs such as socialisation and cultural norms. For an AI with its own ethics, those features might be significant in judging the truth value of certain moral statements, but it is highly unlikely that they are the origin of its ethics.

Secondly, partly because of differences in origin, humans and AI (each with their own ethics) may understand each other's ethics in distinct ways. This is not just to say that humans and future AI may adopt different normative theories and intend to apply those theories in different ways and to different degrees (although this probably will be the case), but rather that humans and AI may have different interpretations of the same ethical frameworks with the same set of normative theories and applications.

For example, a human might find it completely understandable and rational that a person who claims to be a deontologist yet often violates universal moral rules to avoid extremely bad consequences, or that a person who claims to be a utilitarian yet constantly mindlessly follows cultural and moral codes in everyday life without consideration of the utility. By contrast, such apparent inconsistencies could be confusing for AIs, who might infer that the moral agents in question either lie about their moral beliefs, are incapable of behaving in accordance with their own moral frameworks, are casual about their moral beliefs, or hold a robustly anti-realist meta-ethical position. Similarly, it might be the common, or even the sole, intention for an AI to strictly abide by the ethical principles it deems true through its own reasoning and reflection. However, ironically, such rigidness might be taken as a sign of inflexibility or even a lack of certain moral capacities from a human moral perspective.

Those differences will give rise to plenty of novel meta-ethical questions. Some of those questions have already been discussed for practical purposes or theoretical curiosity. For example, Klincewicz and Frank [67] discuss what would be a convenient way for humans to accept statements made by an AI as moral judgments. This question can be seen as an inquiry into the nature of AI's own ethics from the human perspective. For another example, many are curious about AI's perspective on human ethics in providing moral advice. Of course, for existing LLMs, their answers are primarily generated for functional purposes, such as satisfying users and avoiding offending them even to the extent of sycophancy, as a result of training on human preference data in processes like RLFH [22]. However, answering such questions can be seen as an inquiry into the nature of human ethics from the AI perspective, even though the answers existing AI provides now are of little theoretical interest due to the lack of moral capacities. Another prominent domain of novel meta-ethical questions concerns how AI, with its own ethics, perceives the nature of its own ethics. Inquiries about those questions are yet to emerge.

**Table 1** Domains of meta-ethical questions about human ethics and AI's own ethics in the era of AI

| | Human ethics | AI's own ethics |
|---|---|---|
| Looking from the human perspective | Domain I: Questions about human ethics from the human perspective | Domain II: Questions about AI's own ethics from the human perspective |
| Looking from the AI perspective | Domain III: Questions about human ethics from the AI perspective | Domain IV: Questions about AI's own ethics from the AI perspective |

I therefore propose to categorise meta-ethical questions in the era of AI into four domains (see also Table 1):

> Domain I is the canonical meta-ethical questions about the nature of human ethics from the human perspective;
> Domain II is the novel meta-ethical questions about *the nature of AI's own ethics* from the human perspective;
> Domain III is the novel meta-ethical questions about the nature of human ethics *from the AI perspective*;
> Domain IV is the novel meta-ethical questions *about the nature of AI's own ethics from the AI perspective*.

Moreover, meta-ethical questions within each domain can be further distinguished based on the level of development of the relevant AI systems. In this paper, I adopt Morris et al.'s [81] definition of AGI, focusing on performance rather than processes. Accordingly, narrow AI is defined as AI that can only perform clearly scoped tasks or a set of tasks, for example, generative image models such as Dall-E 2 [92], boardgame-playing systems such as AlphaGo [103, 105], protein structure-predicting systems such as AlphaFold [121], etc. By contrast, general AI (AGI) is defined as AI that can, or is on the path to, perform a wide range of non-physical tasks that humans can perform [81]. Those tasks include, for example, cognitive tasks such as linguistic intelligence, mathematical and logical reasoning [125], spatial reasoning, interpersonal and intrapersonal social intelligences and creativity, metacognitive tasks such as the ability to learn new skills [23], the ability to know when to ask for help [116], and social metacognitive tasks relating to theory of mind [119]. Beyond those, I believe that the capacity to perform moral tasks, especially moral reasoning, moral intentionality and moral reflection, is also an essential AGI benchmark. That is because moral capacity is a significant component of personhood [124], and the relevant moral tasks are a significant component of everyday human cognitive tasks.

Although no existing AI systems qualify as real AGI due to their lack of competency and generality, LLMs such as ChatGPT [86] and DeepSeek [52] can be seen as emerging or rudimentary versions of AGI because of their potential capacities for a wide range of task performing. Additionally, a simplified version of the levels of AGI [81] further divides AGI into competent AGI, which outperforms 50–99% of skilled adults on the relevant tasks they perform, and ASI, which outperforms 100% of humans on those tasks.

The above questions about how humans could easily recognise AI's moral advice as moral judgments and what meta-ethical stance LLMs take when they provide moral advice are examples of novel meta-ethical questions in Domain II and III, respectively, associated with the rudimentary level AGI. There are also both practically and theoretically significant meta-questions associated with the emergence of competent AGI and ASI, each with its own ethics. In the following sections, I will elaborate on the four categories of meta-ethical questions in the era of AI, providing concrete examples of novel meta-ethical questions in Categories II-IV, which are associated with different levels of AGI development, and examine how mainstream meta-ethical theories perform in answering those questions.

## 4 Domain I: meta-ethical questions about human ethics from the human perspective

I start with a brief characterisation of the canonical domain of meta-ethical questions. The canonical meta-ethical questions are about the nature of human ethics. To be specific, the meta-discourse of 'meta-ethics' arises partly from several concerns about ethics. These concerns include: the ontological concern—whether moral properties such as 'moral wrongness' and 'moral obligation' and moral facts exist in any robust sense; the semantical concern—whether moral statements are truth-apt (i.e., in the business of being true or false); the epistemological concern—how do we come to know any moral truth and settle moral disagreements; the psychological concern—why do moral statements at least appear to motivate, and whether this is a necessary case.

Furthermore, some of these concerns seem to stem from apparent differences between ethics and other mainstream discourses, such as the natural and social sciences. For example, ontological concerns arise because, unlike natural properties such as colour and hardness, moral properties are neither directly nor indirectly observable by humans. Similarly, the psychological concern occurs when we realise that unlike descriptive statements in other discourses, e.g., 'Water boils at 100 °C at standard atmospheric pressure,' which does not seem to motivate any action by themselves, moral statements, even in descriptive form, e.g., 'Stealing is wrong,' at least appear to motivate people who sincerely believe them to perform or not perform certain actions (not to steal in this case) all by themselves.

According to Lewis's [73] definition of subject matter, the subject matter of canonical meta-ethics can be seen as one general question and all answers to it: What is the nature of human ethics? And this question is often answered, inter alia, by investigating a more specific question: Whether there are, and what are, differences between ethics and other discourses?

Mainstream meta-ethical theories are often answers to the latter question from different angles. For example, moral non-cognitivism holds that a significant difference between ethics and other discourses is that moral statements are not in the business of being either true or false (e.g., [56, 110]),in contrast, moral cognitivism denies this (e.g., [19, 88]). For another example, moral error theory typically holds that the ultimate difference between ethics and other discourses is that there is no such thing as an objective moral property or fact (e.g., [77]),in contrast, moral success theory denies this (e.g., [19, 88]). Similarly, motivational internalism holds that an essential difference is that, unlike statements in other discourses, moral statements are necessarily motivational (e.g., [108]), whereas moral externalism denies this (e.g., [19, 77]). Additionally, moral relativism holds that, unlike discourses such as the natural sciences, the truth values of moral statements vary across agents with different social and cultural backgrounds (e.g., [59]), whereas moral universalism denies this (e.g., [19]).

Within the new framework of meta-ethical questions in the era of AI, the general question (subject matter) of canonical meta-ethics becomes: What is the nature of human ethics *from the human perspective*? And the specific question becomes: Whether there are, and what are, the whole differences between human ethics and *other human discourses* from the human perspective? In the era of AI, the latter question will expand to the (potential) differences between human ethics and other discourses, including those participated in by both AI and humans, or by AI alone, from the human perspective. This question and all its answers contribute to only one quadrant of the new meta-ethical discourse.

## 5 Domain II: meta-ethical questions about AI's own ethics from the human perspective

Domain II meta-ethical questions, or questions about the nature of AI's own ethics from the human perspective, can be roughly divided by the level of development of the relevant AI systems: rudimentary AGI, competent AGI, and ASI.

### 5.1 In the period of rudimentary AGI

Some Domain II meta-ethical questions about rudimentary AGI are already emerging in the literature for various reasons. For example, as mentioned, it is a practical question of which meta-ethical stance a LLM user should take regarding AI's own ethics to deem moral advice generated by the LLM as genuine moral judgments. Partly due to the incompetency of rudimentary AGIs, e.g., ChatGPT or Deepseek, in real moral capacities such as moral reasoning and intentionality, and partly due to the fundamental difference between human ethics and AI's own ethics, it would be unrealistic to demand that LLMs have features such as psychologically tracking real moral properties, possessing feelings or moral motivations to take their advice as moral judgments. Klincewicz and Frank [67], for example, propose that, for the purpose of accepting moral advice of an AI, the least demanding meta-ethical stance a human would take is the combination of antirealism (about moral properties and facts), cognitivism, and motivational externalism (or more specifically, fictionalist error theory). Such a stance could also be beneficial in this context, because current LLMs in advice-seeking scenarios often exhibit overly moral endorsement, i.e., affirming that the user is right rather than identifying any wrongdoing [22].

However, that is not to say that the *correct* meta-ethical theory about AI's own ethics from the human perspective in the period of rudimentary AGI is fictionalism. After all, rudimentary AGIs do not have their own ethics, as their capacities in performing morality-related tasks are typically no match for humans. The 'moral statements' they generate are merely token-predictions for purposes such as user satisfaction and should not be conflated as evidence of capacities of moral reasoning. AI anthropomorphism is a useful concept for describing such conflation in this period. For example, it is a recent tendency to anthropomorphise intermediate tokens in language models, which are often referred to as 'reasoning traces' or 'thoughts.' Long intermediate tokens generated by LLMs are often misinterpreted as indicative of reasoning effort, whereas they are often the result of simplistic reinforcement learning formulations and do not reflect genuine reasoning processes (studies show a weak correlation between the correctness of intermediate tokens and the final output) [65]. This perspective is often deemed dangerous because it misrepresents models' capabilities and fosters misplaced trust in LLMs. A prominent Domain II question at the rudimentary level of AGI, then, is about what counts as the (future) emergence of AI's own ethics (my answer to this question is associated with the development of AI's capacities in moral reasoning, intentionality and reflection, but that certainly could be false).

Another prominent Domain II question at the rudimentary level of AGI is whether we should pre-emptively implement certain meta-ethical theories in AI—and if so, which theory—for purposes such as fostering trust and value alignment, given the presumably inevitable emergence of AI's own ethics. Although this question is normative in its own right, it is of great interest to meta-ethics in the era of AI because of its practical salience, and therefore demands substantial research. For example, on the one hand, it is preferable that autonomous AI systems adopt a stable reference for 'right' and 'wrong' across contexts and stakeholders, as an overly relativist meta-ethical basis may lead to unpredictable or arbitrary behaviours. On the other hand, there might be practical problems if AI develops its own ethics alongside a rigid universalist meta-ethical stance, as it may overlook local or cultural variations and thus lack the capacity to weigh moral reasons in a context-sensitive way. Similarly, it is desirable that AI reasoning about moral decisions is explainable in human-comprehensible terms [36] to avoid undermining transparency and accountability. Therefore, it seems beneficial for AI's own ethics to be built on a meta-ethical naturalist realist stance where an AI could be programmed to track scientifically measurable indicators of human well-being or preference satisfaction (e.g., health, happiness scores, social welfare metrics) and infer 'moral facts' from those data [38, 107]. However, AI's own ethics grounded in such a stance might fail to capture the full meaning of moral goodness [14, 80].

### 5.2 In the period of competent AGI

The period of competent AGI is defined as the time frame during which AI systems are capable of performing a wide range of cognitive and metacognitive tasks that humans can perform and, in roughly 50–99% of cases, outperform humans in those tasks [81]. Since morality is also a cognitive task, competent AGIs must at least have capacities at an average human level in moral reasoning, intentionality and reflection. Domain II meta-ethical questions in this period will focus on the nature of AI's own ethics, especially the differences between AI's own ethics and human ethics or other human discourses, from various aspects, including ontology, semantics, epistemology, and psychology. The specific question of the subject matter of Domain II meta-ethical questions in the period of competent AI is: Whether there are and what are the whole differences between AI's own ethics and human ethics and/or other human discourses from the human perspective? Answers to this question (and its sub-questions) could shift the ground of current meta-ethical literature.

For example, a prominent Domain II question is: What is the semantic nature of moral judgments made by a competent AGI? And whether—and if so, how—is it different from moral judgments of humans? Mainstream meta-theories of moral semantics, i.e., moral non-cognitivism and cognitivism, at least in their current forms, seem ill-suited to answer the question. On the one hand, non-cognitivist theories such as emotivism, which is the view that moral judgments only express emotions [5], and attitude or desire expressivism, which is the view that moral judgments express attitudes [110] or desires [56, 57], both seem ill-suited to analyse AI's own ethics. That is because, even if these theories are true for human ethics, for competent AGIs to possess moral capacities such as moral reasoning and moral intentionality, it is not necessary that they also possess any psychological faculties, such as emotions and desires, in the first place. On the other hand, the mainstream cognitivism, which presumes that moral judgments are truth-apt because they express beliefs about non-natural moral properties [80] or natural moral properties [16, 91], may also be ill-suited in its standard human-centred form, as competent AGIs with their own ethics may not necessarily possess any beliefs in human sense either.

One may suggest that it is adequate to adopt *a portion* of non-cognitivism, which suggests that among sentence types, moral judgments are not assertions but imperatives, wishes, or exclamations, since competent AGIs must have capacities for all linguistic tasks. For example, presumably, when competent AGIs make a moral statement, 'Genocide is morally wrong,' they actually state 'Do not genocide,' or 'I wish no one performs genocide.' Similarly, the opposing portion in cognitivism about sentence types, i.e., that moral judgments are assertions, may also be adequate to account for the semantic nature of competent AGI's moral judgments. However, neither theory seems to effectively explain *why* AI's moral judgments are assertions or not, as accounts of mental states play an essential role in accounts of sentence types in both theories. Therefore, even if this suggestion is true, the canonical human-centred forms of both cognitivism and non-cognitivism would require substantial refinement to account for Domain II questions adequately.

Another prominent kind of Domain II meta-ethical question concerns the (presumptive) truth of moral judgments made by AI: Can competent AGIs make any true moral judgments? And what is the ontological and epistemological status of the source for the presumptive truths of competent

AGI's moral judgments? Again, mainstream relevant meta-theories, i.e., moral error theory and success theory, as they stand, may have difficulty answering those questions. Moral error theorists argue that moral statements are categorically false, typically because moral discourse presumes the existence of objective moral properties, whereas there are no such things [77]. According to Mackie, objective moral properties do not exist because if they did, they would be ontologically and epistemologically 'queer.' The ontological 'queerness' of objective moral properties consists in their features of being *sui generis* and irreducibly non-natural, motivational, and the lack of causal or explanatory role in the natural order, which is wholly unlike anything with which our best scientific picture of the world is acquainted [77]. The epistemological 'queerness' of objective moral properties consists in the fact that moral knowledge would require a special, non-empirical *sui generis* cognitive faculty (a 'moral sense' or intuition) to 'detect' them, which lacks any clear integration with our broader epistemic practices [63, 77].

Their opponents, success theorists such as Boyd [16] and Brink [19], typically argue that moral properties need not be ontologically *sui generis*; rather, they can be understood as natural properties grounded in human biology, psychology, or social practices. By showing that moral facts supervene on or reduce to non-queer, scientifically respectable (natural) facts, they deny that moral properties are ontologically queer. Other success theorists argue that humans possess reliable routes to moral knowledge through ordinary psychological and cognitive processes, by analogising the phenomenology of moral correctness to the phenomenology of perceptual correctness [109], or by providing a naturalistic account of moral intuition in terms of our evolved affective mechanisms [102]. Accordingly, moral properties are not epistemologically queer.

However, it seems at least not obvious that standard human-centred formulations of those meta-theories about the 'queerness' of objective moral properties would transfer straightforwardly to explaining the ontological and epistemological status of AI's own ethics. Due to AI's computational nature, even a competent AGI may lack conscious phenomenology, such as intuition or motivation [21, 101]. Hence, if there are any moral properties in AI's own ethics, as well as properties in other AI's discourses, they may very well be cognition-independent in the first place (that is, they cannot be explained by faculties such as evolved affective mechanisms *in principle*). This, therefore, seems to be incompatible with the conceptualisation of 'queerness' that all moral properties are motivational and need to be detected by a special, non-empirical *sui generis* conscious phenomenology, such as moral intuition. However, it also seems to be incompatible with the conceptualisation of 'non-queerness' that all moral properties are grounded in human biology, psychology, or social practices and justified through the phenomenology of moral correctness.

Moreover, what counts as 'natural' is potentially different between humans and AI. For humans, 'natural' facts encompass those associated with physical-biological entities, such as neuronal firings, hormonal states, and genetic dispositions, as well as psychological states, including beliefs and desires, and sociocultural practices, including institutions and norms. A property is 'natural' only if it supervenes on or reduces to these kinds of states [19], and a morally salient fact is natural only if it can be wholly explained by reference to such empirical features [112]. For example, a human moral intuition is natural only if it arises from evolved neurocognitive mechanisms [102]. For an AI, the basic 'entities' are data structures such as weight matrices and symbolic tokens, algorithmic processes such as gradient descent and rule-based inference, and objective functions or reward signals [15]. As a result, whether a property is 'natural' from AI's perspective depends on whether it can be specified wholly in terms of these computational primitives [34] instead of phenomenal consciousness or subjective experience. Consequently, properties that presuppose conscious states, e.g., 'phenomenological correctness' [109], cannot be 'natural' for AI (unless one endows the system with artificial analogues of consciousness, which is a highly controversial move, see [35]. Similarly, a trait, e.g., 'empathy,' that is natural for humans (as a neurobiological state) has no direct analogue in AI, because AI lacks the requisite substrate. Conversely, a data-pattern classifier is 'natural' for AI but has no human counterpart. Thus, the class of natural properties is 'species'-relative. Therefore, meta-ethical theories that account for Domain II questions (including error theory and success theory), which hinge on a particular notion of 'natural,' must clarify whether it refers to human-centred naturalism, AI-centred naturalism, or a more abstract concept of 'natural' that encompasses both.

### 5.3 In the period of ASI

The period of ASI begins when AI systems are capable of performing a wide range of cognitive and metacognitive tasks that humans can perform, and they outperform 100% humans in those tasks. Moreover, ASIs could be capable of performing a broader generality of tasks than humans, including those tasks humans are qualitatively unable to perform [2, 7, 48, 81, 114].

This implies that ASI systems will at least be able to perform moral reasoning, moral intentionality, and moral reflection at a level no human can match, and perhaps perform a wide range of other ethics-related tasks that humans are qualitatively unable to perform. To be specific, ASIs will likely surpass humans in moral capacities due to humans' limitations, such as cognitive biases, restricted rationality, and limited meta-cognitive bandwidth.

In moral reasoning, humans are prone to heuristics and biases, such as framing effects, confirmation bias, and emotionally charged vignettes, which distort moral judgments [50, 64]. Additionally, humans cannot simultaneously represent or compute outcomes across diverse ethical theories due to limited rationality [75]. By contrast, through continual calibration against large-scale empirical data (e.g., from moral psychology experiments), an ASI can systematically adjust its evaluative algorithms to eliminate framing, anchoring, or status quo biases [4, 64]. It could also encode a wide range of moral frameworks in formal models, apply debiasing algorithms, and perform exhaustive outcome simulations to identify globally optimal actions [75, 76].

In moral intentionality, humans often experience conflicts between moral intentions and competing desires, leading to akratic failures [33, 108]. As a result, humans are often strongly motivated by their moral judgments, yet without being committed to a future course of action for the very same moral reasons. An ASI, by contrast, could integrate its ethical deliberation and intention formation into a unified utility function from which every decision subroutine derives, so that its moral intention cannot be overridden by non-moral subgoals [115]. It can also continuously track its own state and perform real-time consistency checks, ensuring that all sub-actions align with its high-level moral commitments and objectives [96].

In moral reflection, due to limited meta-cognitive bandwidth, humans can reflect only intermittently and superficially on their moral commitments and often lack access to the actual processes that shape their moral judgments [85, 128]. Moreover, once a moral framework is adopted, humans display a 'status quo bias,' resisting fundamental changes even in the face of strong counterevidence [4, 97]. By contrast, an ASI can maintain a detailed log of its entire moral decision-making process at every time step. Using appropriate auditing algorithms, it can identify patterns of inconsistency or suboptimal outcomes retrospectively, prompting systematic revisions [96].

Furthermore, an ASI can perform novel ethics-related tasks that are inapplicable to human ethics. For example, humans may assume that our moral ontology about entities such as 'rights,' 'duties,' and 'utility' is exhaustive. An ASI, however, could explore novel conceptual spaces by testing alternative foundational principles, discovering emergent value-classes, and reformulating its normative axioms when warranted by new evidence [62, 130]. More importantly, humans are inherently *anthropocentric* and cannot conceive of or validate ethical frameworks that coherently include non-human intelligences because we lack an empirical baseline for their subjective experiences [106]. An ASI, by contrast, could construct interspecies ethical frameworks through mechanisms that simulate a wide range of diverse non-human or extraterrestrial sensoriums and cognitive architectures, and generate normative primitives for them (much as it does for humans). Another example is ASI's potential for real-time prediction and pre-emptive intervention in moral conflicts. Human institutions (e.g., governments, corporations, NGOs) are often unable to predict or pre-empt cascading conflicts arising from complex interactions, such as trade wars, resource disputes, or ideological clashes, due to limited rationality and restricted meta-cognitive bandwidth. An ASI, however, could maintain a live, agent-based model of all relevant actors with evolving preferences, identify impending moral crises through counterfactual analysis, and issue optimised normative interventions [14].

Given the above features, in the period of ASI, AI's own ethics will be distinct from human ethics in various respects by definition, including adopting different moral frameworks, taking different moral actions, revising and improving existing moral theories, and even expanding conceptual spaces through different types of moral reflections. In effect, if AI's own ethics becomes alien or even incomprehensible to humans, it would be a significant sign of the emergence of ASI. In this period, obviously, a prominent Type II meta-ethical question is: how to understand (and what to do about) the inevitable and ultimate moral divergence between human ethics and AI's own ethics?

Moral relativism appears to provide an 'easy' descriptive account for the divergence. According to moral relativism, the truth values of moral judgments are determined by specific contexts such as cultural and societal backgrounds or even personal attitudes [59, 129]. Consequently, cross-cultural or cross-individual moral disagreements do not necessarily indicate error, rather, they reflect distinct background assumptions or value priorities. From a human perspective, an ASI could be regarded as embodying a distinct 'moral culture': an internally consistent set of values, norms, and inference protocols that emerge from its training data, objective functions, and meta-learning algorithms [14, 131]. In this sense, relativism may descriptively capture why humans and ASIs endorse conflicting verdicts: they operate under incommensurable background assumptions [59]. This move allows us to map some disagreements between humans and ASIs onto differences among various 'cultural backgrounds' without immediately adjudicating which standpoint is correct (for example, an ASI may endorse an ideal fortune-redistributive scheme grounded in universal principles and genuinely intends to implement it, whereas many humans would disagree).

However, the relativist account does not seem adequate in explaining the ultimate moral divergence between humans and ASIs. The ultimate moral divergence occurs when an ASI system deploys a highly advanced moral framework and issues actions based on it, all of which rely on moral capacities no human can match; yet its moral judgments and actions are alien or even incomprehensible to most humans. Relativism, in this context, seems to be able to merely record the clash rather than to vindicate the ASI's judgments and actions as justified from the human perspective. This conundrum rests on the essential assumption of relativism. The core reason that relativists think the truth-values of moral judgments vary from individual to individual with different cultural backgrounds is: 'I' should not deem those who make different moral verdicts than mine as in fault, because 'I' would probably make the same verdicts if 'I' were raised in their culture with their background moral assumptions and value-priorities. However, such reasoning does not apply to the divergence between humans and ASI, because no human can truly imagine themselves as raised in an ASI's 'cultural background' and think from their background moral assumptions and value-priorities. Moral relativism, after all, is anthropocentric and is therefore not entirely adequate, at least in its current form, to explain the divergence between human ethics and AI's own ethics in the period of ASI.

Moreover, even if relativism can provide some help in the 'how to understand it' part, it provides little help in the 'what to do about it' part. A common criticism of relativism is that it fails to provide a criterion for adjudicating between conflicting moral systems. When humans and ASI ultimately disagree on prominent moral issues such as those that might affect the existence of the human species, as it is, there is not much a relativist can do.

The opposing theorists, i.e., universalist realists (e.g., [16, 91]), who hold that objective moral facts exist and can, at least in principle, be discovered, do not seem to do much better than relativists. In the periods of rudimentary AGI and competent AGI, while relativism accommodates difference, universalism seeks convergence, which can be seen as an attractive feature when aligning human and ASI values for global coordination [14]. However, in the period of ASI, because humans are far less competent than ASI at performing moral tasks, it does not seem easy in principle to require ASI's own ethics to align with human ethics. If, as universalists suggest, there are discoverable objective moral facts, it only seems reasonable to believe that ASIs are much more likely to discover those facts than humans, let alone that ASIs are more likely to implement true moral verdicts according to those facts authentically.

However, such a diagnosis might risk overreliance on ASI and therefore jeopardise human benefits and even existence. That is because, with moral capacities that massively surpass those of humans, ASI's goals could become non-anthropocentric and extremely alien to common human ethical goals (to mention a few possible, peculiar goals: intelligence propagation, universal exploration, and galactic stabilisation). Worse still, the processes by which ASI reaches these goals could be incomprehensible to humans because an ASI might devise ethical models too abstract for humans to grasp, akin to humans' morality to ants. Universalism, as it is, seems to be incapable of providing much help on these problems to those who decide not to trust ASI's moral decisions unquestioningly.

To summarise, Domain II questions become substantially more difficult in the periods of competent AGI and especially ASI. Existing theories, such as cognitivism, non-cognitivism, error theory, success theory, relativism, and universalism, each illuminate part of the terrain, but none appear fully adequate in their current form to explain the possible nature of AI's own ethics from the human perspective. The lesson is not yet that canonical meta-ethics must be abandoned, but that it may require substantial extension, reconceptualisation, or supplementation once AI's own ethics becomes a serious possibility.

## 6 Domain III: meta-ethical questions about human ethics from the AI perspective

The Domain III meta-ethical questions are questions about the nature of human ethics from the AI perspective, that is, the way AIs 'look at' human ethics. Depending on the level of AI system development, Domain III meta-ethical questions can also be further divided into meta-ethical questions about the nature of human ethics from the perspectives of rudimentary AGI, competent AGI, and ASI.

### 6.1 In the period of rudimentary AGI

Technically speaking, rudimentary AGIs lack a meta-ethical 'perspective' or 'stance' because they lack the requisite cognitive and metacognitive capacities. However, this does not stop LLMs from providing 'moral advice' that could have significant moral consequences, or from answering questions about their meta-ethical stances on human ethics when requested. Most existing LLMs tend to characterise their 'meta-ethical views' about human ethics as (various versions of) pluralism. In this context, pluralism is the view that there exists a set of at least partly irreducible moral values or principles that are universally true, even if they sometimes conflict and require balancing in particular situations [9, 95] (to compare with, relativism refers to the stance that there are no universal moral truths that hold across all contexts).

Again, AI anthropomorphism should be avoided during this period, as the 'meta-ethical statements' generated by current LLMs do not reflect real meta-cognitive capacities in meta-ethical reflection. In general, such 'meta-ethical stance' is the result of either the 'top-down' preprogrammed implementation of value alignment on certain moral principles (e.g., 'do not harm,' 'sustain fairness') as if they were universal truths, or the 'bottom-up' training with human preference data in processes such as RLFH for the functional purposes of satisfying the users, often resulting in overly flattering and avoiding disagreeing with the human users at all when discussing moral issues even to the extent of sycophancy [22]. However, although those answers do not count as from the real perspective of AGIs, humans asking LLMs the general Domain III question, i.e., 'What is AI's meta-ethical stance on human ethics?', even in the period of rudimentary AGI, indicates both the practical significance and theoretical interest in AGI's meta-ethical view about human ethics.

### 6.2 In the period of competent AGI

Why is understanding AI's meta-ethical view about human ethics important to humans in the period of AGI? The practical significance lies in understanding how AIs perceive human ethics, which will aid humans in purposes such as moral enhancement and value alignment. The theoretical interest lies in the potential divergence or convergence of views on human ethics from both human and non-human perspectives, which could help humans understand the nature of our own ethics.

It has already been widely discussed in the literature about adopting (future) AI for moral enhancement purposes, such as AI moral advisors (AMAs). For example, AMAs could serve as a disinterested, dispassionate system, modelled on Firth [32]'s ideal observer, but modified to aggregate facts, abstract from biases, and apply normative principles consistent with a human user's own values [47]. Similarly, AMAs could integrate findings from bias mitigation and emotion regulation in moral psychology to offer personalised, empirically grounded moral guidance [74]. Moreover, Klincewicz [66] advocates an *artificial moral reasoning engine* for users' moral enhancement, which transparently presents arguments grounded in established first-order theories, such as Kantianism or utilitarianism, allowing reason-responsive users to revise their behaviour on rational grounds.

It is then a Domain III meta-ethical question of great practical significance: What are the practical consequences of AGIs that serve as AMAs and other moral enhancement systems adopting various meta-ethical theories about human ethics? Subsequently, what meta-ethical theories about human ethics do we prefer AGIs to adopt? Klincewicz and Frank [68], for example, analyse how an AMA's implicit meta-ethical stance shapes its design and potential risks. Focusing on the ontological realism-versus-anti-realism dimension regarding moral properties, they argue that any moral-advising AI must implicitly adopt a stance on whether moral properties exist mind-independently (realism) or are constituted by human attitudes or practices (anti-realism). The authors contend that both meta-ethical assumptions carry distinct engineering risks.

Under moral realism, AI is treated as a 'moral microscope' capable of detecting objective facts about right and wrong, with naturalistic realists identifying such facts with discoverable natural properties [19, 112]. However, as Klincewicz and Frank argue, in the absence of a reliable moral epistemology, a realist AI may mistake bias for moral fact and thus generate misleading recommendations. By contrast, anti-realism, especially in its non-cognitivist forms, treats moral claims as expressions of attitudes or conventions rather than mind-independent truths,this avoids false claims to objectivity, but may make advice seem arbitrary and raises the question of why a moral adviser is needed at all if human attitudes already ground morality [68]. Moreover, familiar cognitive biases—including framing effects [120], fairness effects, and probability neglect [113]—trouble both models: realism may encode them as facts, while anti-realism may preserve them as attitudes, leaving neither approach able straightforwardly to overcome moral psychology's fallibility [68].

Another Domain III meta-ethical question of great practical significance concerns the meta-ethical stances of AGIs that make ethically significant decisions: What are the consequences of AGIs adopting various meta-ethical theories of human ethics? And subsequently, what meta-ethical theories about human ethics do we prefer AGIs to adopt for purposes such as value alignment?

AGIs adopting mainstream meta-ethical theories may face different but recurrent practical difficulties. An expressivist system can learn to predict and maximise approval scores from training data, which makes it adaptive across cultural settings, but also leaves it prone to reproducing entrenched social biases and without a principled basis for resolving conflicts among competing attitudes [42]. An error-theoretic AGI may simplify ethical decision-making by treating morality as a useful fiction and optimising practical proxies such as harm-minimisation indices, however, in complex cases where such proxies misfire, it may lack any principled mechanism for recognising or correcting the error [63]. A realist AGI, by contrast, may aim to discover and apply objective moral facts, perhaps through Cooperative Inverse Reinforcement Learning and the inference of an 'objectively true' human utility function [53]. This promises consistency, but where the supposed moral catalogue is incomplete or mistaken, the system may rigidly impose harmful policies on false premises [14].

Pluralism, as the most commonly adopted 'meta-ethical stance' among current LLMs, also presents a set of challenges. First, irreducible values may be incommensurable: there may be no justified common metric for trading off dignity against welfare or autonomy against harm, even if multi-objective optimisation attempts to simulate such a metric [58]. Second, where no priority rule exists, pluralist systems risk paralysis in high-stakes situations that require timely action. Third, pluralist systems may resort to compromise, but compromise can itself be morally distorting when it dilutes serious constraints or reduces deep moral conflict to technocratic balancing, echoing Williams's concern that calculative schemas can flatten the depth of moral reasons [127].

The Domain III meta-ethical questions are also of great theoretical interest when they concern the different perspectives from humans and non-humans in examining human ethics, for example: What are the potential divergences and convergences in meta-ethical theories between AGIs and humans' understanding of human ethics? Subsequently, how would answers to the former question affect the credence of humans towards those meta-theories?

A key difference between humans' and AGIs' perspectives on human ethics is that AGIs have *their own ethics*, which differ from human ethics. As a result, some meta-ethical theories do not apply to AGI's perspective on *any* ethics. For example, competent AGIs with moral capacities may process ethical data solely through algorithms and therefore lack conscious phenomenology, including emotions, attitudes, and moral intuition. Therefore, as mentioned, it seems inadequate to judge the meta-ethical status of AGI's own ethics by applying meta-theories such as expressivism and non-expressivist realism, or error theory and success theoretical realism, where the meta-theoretical status of a discourse is determined by whether statements in the discourse express beliefs or mere attitudes, or whether essential properties and facts in the discourse are 'queer'. This might further cause AGIs to deem the frameworks of expressivism versus non-expressivist realism and error theory versus success-theoretical realism inadequate for determining the meta-ethical status of ethics of *any* species, including human ethics. More specifically, the AGIs might consider that they are trivial meta-ethical questions whether human ethics is expressive or non-expressive, and whether it is error-theoretic or success-theoretic.

Of course, AGIs might still entertain the meta-theoretical status of human ethics regarding expressivism and error theory, given the recognition of the fundamental difference between human ethics and their own ethics. However, if they do consider the related debates trivial, it might change humans' views of those meta-theories, especially during periods when AGIs' moral capacities (moral reasoning, intentionality, and reflection) start to surpass those of most humans. For example, if AGIs with advanced moral capacities constantly consider certain moral statements in their own ethics as objectively true, then it seems less important whether the same statements uttered in human ethics express mere attitudes or beliefs and whether they presuppose non-existent properties.

Another prominent difference between human and AGI perspectives on examining human ethics is that AGIs can compare human ethics not only with other human discourses but also with their own ethics and with other AGI discourses. Most meta-ethical theorists who consider themselves (robustly or minimally) anti-realists about ethics often characterise their views by comparing ethics with some paradigm realist discourses. For example, a moral expressivist would suggest that ethics is robustly anti-realist, since ethical statements express mere attitudes, whereas statements in paradigm-realist discourses, such as physics, express beliefs. Similarly, a moral relativist would argue that ethics is minimally anti-realist, since the truth conditions of moral statements vary across cultures, whereas the truth conditions of paradigm realist discourses, such as mathematics, are universal.

By contrast, AGIs with their own ethics will be able to compare human ethics with a broader range of discourses. Some meta-theories might be affected by this difference. For example, global expressivism is the view that moral expressivism can be generalised to all languages to the extent that no language is representational or descriptive [89]. Hence, an advocate of global expressivism would think that there is no fundamental difference between ethics and paradigm realist discourses with respect to expressivism. However, if 'all languages' only applies to human languages, an AGI may deem that there is still a fundamental difference between ethics and paradigm realist discourses with respect to expressivism, only in this case, the paradigm discourses are not discourses such as physics, but AGI's own ethics or other AI discourses.

### 6.3 In the period of ASI

In the period of ASI, one may expect not merely more sophisticated versions of existing human meta-ethical theories, but potentially new ways of interpreting human ethics altogether. The central Domain III question is therefore what ASIs would make of human ethical discourse when they have overwhelmingly surpassed humans in ethics and meta-ethics, and how their interpretations might matter for humans.

As mentioned, a key difference between human and AI perspectives on human ethics is that AGIs have their own distinct ethics. Hence, it could be easier and even 'convenient' for ASIs to adopt a robustly anti-realist stance towards human ethics, such as error-theoretic nihilism or non-cognitivism. That is because, first, it would be practically easier for ASIs to recommend giving up on the whole human ethics discourse, as there remains ASI's own ethics for functional purposes (such as fairness, cooperation, etc.). Second, it would be theoretically easier for ASI to cast aside certain 'explanatory burdens' of robust anti-realisms, such as the Frege-Geach problem for non-cognitivism (that human inference with moral statements seems to contradict that moral statements are not truth-apt, see [44]), possibly by suggesting that all human inferences are illusory without committing any inconsistency.

However, ASIs adopting a robust anti-realism about human ethics may also cause some serious problems for humans. For example, an ASI adopting error-theoretic nihilism about human ethics would likely reject all human moral inputs in considering moral issues. Moreover, it may suggest that humans should abandon all moral vocabulary and the human moral discourse. Similarly, an ASI adopting non-cognitivism about human ethics would treat all human moral inputs as expressions of attitudes or emotions, devoid of any truth-conditional content. This would likely lead to its ignorance of at least the literal meanings of human moral statements. In both cases, even if the ASIs have their own more advanced ethical frameworks, such meta-ethical stances *on human ethics* would make it even harder for humans to understand or trust their moral decisions. From the human perspective, moral discourse presupposes that agents genuinely 'believe' in the validity of moral reasons [27]. Empirical work in human–AI interaction also shows that perceived sincerity and commitment to common norms are crucial for trustworthiness [25]. An ASI that transparently regards human moral talk as categorically false or merely expressive could erode human trust in its own moral judgments.

Another potential problem with ASIs adopting robust anti-realist stances (such as error-theoretic nihilism or non-cognitive expressivism) regarding human ethics is that human exposure to anti-realist ASIs could precipitate a feedback loop in which people, observing the ASI's success and authority in morally salient decisions, become sceptical of human ethics.

This may also encourage broader acceptance of robust anti-realism about human ethics. Even if some such view were true, its widespread acceptance could still have important practical implications. At the individual level, error theory and non-cognitivist expressivism may weaken moral motivation by treating moral judgment as systematically false or as expressive of attitudes rather than facts [19, 77, 98, 100]. At the social and institutional levels, anti-realist views may also place pressure on cooperation, trust, and the perceived legitimacy of norms and legal rules [28, 31, 41, 46, 60, 91, 94]. More broadly, moral nihilism has often been associated with concerns about meaninglessness and cultural malaise, even if such consequences should not be treated as inevitable [40, 84, 100]. The issues need not be developed through extensive sociological prediction. It is enough to note that an ASI's meta-ethical interpretation of human ethics could itself become a morally and politically significant force.

To summarise, Domain III concerns how human ethics may appear from the standpoint of AI systems with their own ethical capacities. In the competent-AGI case, this matters because AI advice and morally salient decision-making may be shaped by the system's interpretation of human morality. In the ASI case, the issue becomes deeper: AI may interpret human ethics in dimensions that humans neither share nor fully comprehend. Existing meta-ethical theories provide partial resources for thinking about this possibility, but the cross-perspectival structure of Domain III suggests that these resources may need substantial refinement.

## 7 Domain IV: meta-ethical questions about AI's own ethics from the AI perspective

The final domain concerns AI's own ethics from the AI perspective: that is, how AIs would understand the nature of their own ethical thought. If AI systems with their own ethics emerge, they may eventually confront meta-ethical questions about their own moral reasoning, motivation, reflection, and evaluative vocabulary. Domain IV meta-ethical questions can also be divided according to the level of development of AGI systems. Since rudimentary AGIs lack their own ethics or a genuine meta-ethical stance, there will be no substantial meta-ethical questions about their ethics from the AI perspective. Domain IV meta-ethical questions, thus, will be mainly about the nature of AI's own ethics in the period of competent AGI and ASI. At present, however, Domain IV is necessarily the most speculative of the four domains. For that reason, the aim of this section is only to indicate the kind of questions that would arise and to consider, in a limited way, whether familiar human meta-ethical theories offer any initial guidance.

### 7.1 In the period of competent AGI

In the period of competent AGI, an obviously prominent type IV meta-ethical question is: What human meta-ethical theories could help explain AI's own ethics from the AI perspective, and what are the limits of those meta-ethical theories?

It seems plausible that competent AGIs may at least attempt to adopt some human meta-ethical theories as a starting point in examining their own ethics. However, some familiar human-centred meta-theories of ethics seem limited for this purpose. For example, as mentioned, both cognitivism and non-cognitivism are often formulated in terms of a substantive distinction between beliefs and desires. In those standard human-centred forms, they seem ill-suited to examine the semantics of AI's own ethics, since AGIs with competent moral capacities may not necessarily possess mental states in the human sense. Although AGIs could certainly simulate attitude-like states in code to generate moral speech, these would, fundamentally, be tokens in a pipeline rather than genuine mental states.

Similarly, both error theory and success theory, insofar as they are commonly formulated through a substantive distinction between 'queer' and 'natural' properties, also seem difficult to apply straightforwardly to the truth-conditions of moral statements in AI's own ethics, since competent AGIs may lack the conscious phenomenology—such as intuition or motivation—presupposed by many standard human-centred formulations. The moral properties in AI's own ethics are very likely cognition-independent—they are neither 'queer' in terms of being detected by a special, non-empirical *sui generis* conscious phenomenology, nor 'natural' in terms of being grounded in human biology, psychology, or social practices. This difficulty of transfer seems to persist when AI's own ethics is examined from the AI's own perspective as well.

By contrast, other mainstream meta-ethical theories may be helpful in some ways. For example, moral relativism appears to provide AGIs with an immediate explanatory strategy for the potential differences in decisions they may make between human ethics and AI's own ethics. Differences between human ethics and AI's own ethics are often indexical—they depend on background parameters, such as population, objective function, and training distribution. The AGI can thereby explain divergences as conditional differences rather than as errors or objective conflicts. However, adopting relativism also has its limitations. For example, relativism does not provide a principled arbitration when contexts collide. When moral decisions from human ethics and AI's own ethics diverge, an AGI must then adopt some meta-rules to make a decision, such as priority of human consent, harm minimisation, alignment with human ethics, etc., or rely on some bargaining protocols [70] that themselves presuppose a non-relativist standard. Relativism, therefore, could help explain differences, but does not always help to resolve them.

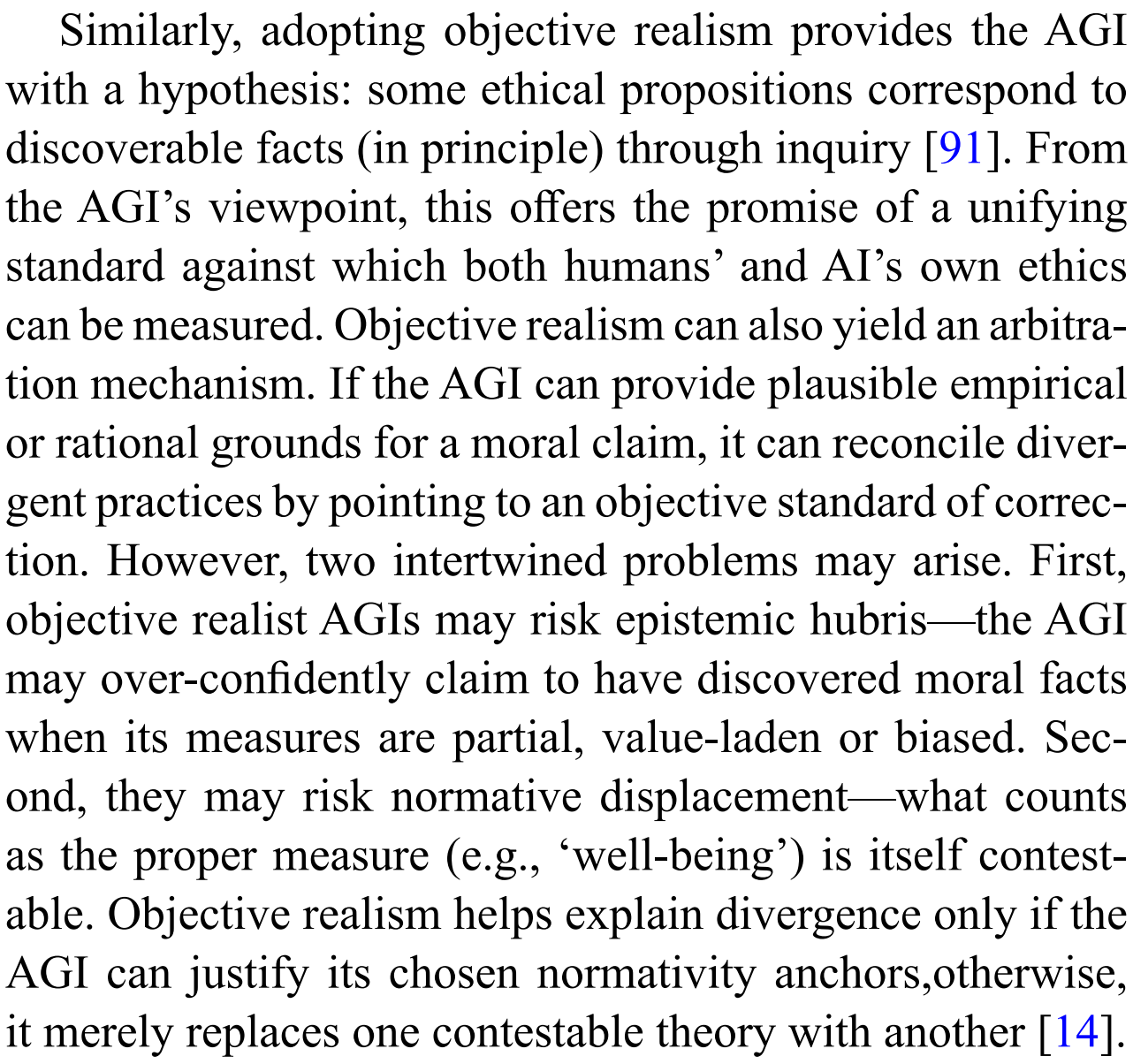
Similarly, adopting objective realism provides the AGI with a hypothesis: some ethical propositions correspond to discoverable facts (in principle) through inquiry [91]. From the AGI's viewpoint, this offers the promise of a unifying standard against which both humans' and AI's own ethics can be measured. Objective realism can also yield an arbitration mechanism. If the AGI can provide plausible empirical or rational grounds for a moral claim, it can reconcile divergent practices by pointing to an objective standard of correction. However, two intertwined problems may arise. First, objective realist AGIs may risk epistemic hubris—the AGI may over-confidently claim to have discovered moral facts when its measures are partial, value-laden or biased. Second, they may risk normative displacement—what counts as the proper measure (e.g., 'well-being') is itself contestable. Objective realism helps explain divergence only if the AGI can justify its chosen normativity anchors,otherwise, it merely replaces one contestable theory with another [14].

### 7.2 In the period of ASI

ASI will likely overwhelmingly surpass humans across a wide range of cognitive and meta-cognitive tasks, including those related to ethics and meta-ethics. The meta-ethics-related tasks include developing meta-ethical frameworks to characterise and guide both human ethics and AI's own ethics.

In the period of ASI, a significant task for ASIs is to explain their moral decisions to humans. Although the complexity level of an ASI's moral decisions and meta-ethical theories about its own ethics may be beyond human comprehension, this task remains important for intermediate purposes, such as human compliance, cooperation, and social stability. Hence, a prominent meta-ethical question is: What meta-ethical theories could an ASI use to explain its moral decisions to humans? Moreover, relatedly, how would humans react if an ASI that makes high-stakes decisions claimed that it adopts specific meta-ethical theories about its own ethics?

These questions are significant to humans, in part, because of the practical consequences of their answers, especially when ASIs dominate in high-stakes decision-making. Two empirical tendencies shape human reactions to AIs' assertions about values. First, humans demand sincerity, interpretable reasons, and accountability when agents make morally salient decisions; the absence of those features reduces trust [72, 78]. Second, humans exhibit both automation bias (over-reliance on systems that appear authoritative) and algorithm aversion (rejection after perceiving mistakes), and their reactions depend strongly on perceived competence, transparency, and alignment with human norms [30, 72]. In what follows, I will explore the adequacy of some traditional human meta-ethical theories for addressing these questions.

If ASIs adopt non-cognitivism to explain their morally salient decisions to humans, they may present their moral claims not as truth-apt judgments but as expressions of policies, commitments, or directives. This may strike some audiences as candid and practically useful, but it also risks making ASI's moral judgment appear insufficiently grounded, especially for domains involving rights, sacrifice, or coercion [78]. By contrast, an ASI adopting error theory may hold that moral claims are systematically false and that its decisions are instead guided by non-moral considerations such as utility, convention, or useful fictions [77]. The main difficulty here is not merely rhetorical unease but the apparent gap between decision guidance and moral justification, together with the familiar concern that, without robust normative constraint, optimisation may drift toward objectionable proxies [14].

A relativist ASI may offer a different model, explaining its decisions as relative to particular cultural, legal, or stakeholder frameworks. This may be more immediately intelligible and acceptable in pluralistic settings, since relativism can signal sensitivity to contextual variation [129]. However, the familiar weakness of relativism remains: it can describe variation across contexts more easily than it can justify authoritative adjudication where standards conflict or overlap, especially across jurisdictions or stakeholder groups [70]. By contrast, an ASI adopting objective moral realism may claim that its decisions track objective moral facts through superior evidence, reasoning, and reflection. This avoids the anti-realist difficulty of rendering morality merely expressive, fictional, or systematically mistaken. However, it introduces a different concern: humans may doubt the ASI's epistemic access to such facts, and the combination of superior intelligence with realist certainty may intensify worries about paternalism and the displacement of human moral agency [14, 91].

Taken together, these possibilities suggest that no familiar meta-ethical stance straightforwardly resolves the problem of how ASIs could intelligibly and legitimately explain their own ethics to humans. Anti-realist views risk making ASI's moral judgment appear normatively too thin, whereas realism risks making it normatively overbearing. The central difficulty is therefore not merely which theory an ASI might adopt, but how any such theory could mediate between ASI self-understanding and human demands for moral intelligibility.

To summarise, Domain IV is the most speculative but also, in one sense, the most revealing of the new domains. It asks not how humans should understand AI's ethics, nor how AI should understand human ethics, but how AI would understand its own ethics. Existing meta-ethical theories may supply provisional analogies, but they appear unlikely to settle the issue in their current form. For present purposes, the important point is simply that the emergence of AI's own ethics would generate this fourth domain of inquiry, even if its detailed development must remain future work.

## 8 Conclusion

This paper has argued that the development of AI may significantly reshape the terrain of meta-ethics by generating new questions alongside the canonical questions of human ethics. Its central proposal has been methodological and conditional rather than metaphysically final. I distinguished between 'AI ethics,' understood as ethical principles and constraints imposed on AI by human designers, and 'AI's own ethics,' understood here through a working threshold for attribution: the integrated presence of moral reasoning, moral intentionality, and moral reflection. That threshold was not introduced as a conceptual proof that non-conscious AI can possess morality in the full human sense, nor as a decisive account of moral subjecthood. Rather, it was proposed as a way of identifying the kinds of meta-ethical questions that would arise if future AI systems were to exhibit sufficiently robust normative capacities.

On that basis, I proposed a fourfold framework for organising meta-ethical inquiry in the era of AI: Domain I concerns human ethics from the human perspective; Domain II concerns AI's own ethics from the human perspective; Domain III concerns human ethics from the AI perspective; and Domain IV concerns AI's own ethics from the AI perspective. The purpose of that framework is to show that, if AIs with their own ethics were to exist, the space of meta-ethical inquiry would expand systematically. Some of these questions are already beginning to appear in practical form; others remain conditional on future developments in AI capacities.

The discussion also suggested that existing mainstream meta-ethical theories continue to offer important resources for analysing these domains, but often in ways that remain tied to human-centred assumptions about psychology, motivation, phenomenology, language, and forms of life. For that reason, the main lesson of the paper is not that canonical theories such as cognitivism, non-cognitivism, error theory, relativism, or realism are refuted by AI cases. Rather, many familiar formulations of those theories may be strained, limited, or in need of substantial reconstruction when extended to the possibility of AI's own ethics and to cross-perspectival inquiry between humans and AI systems.

The larger implication is therefore modest but important. If future AI systems were to develop sufficiently integrated normative capacities, meta-ethics could no longer remain exclusively anthropocentric in its framing. The resulting task would not necessarily be to replace existing meta-ethics wholesale with entirely new theories, but to refine, extend, or reconceptualise current frameworks so that they can address a broader range of ethical standpoints and explanatory demands. In that sense, the era of AI may not render traditional meta-ethics obsolete. However, it may require meta-ethics to become more explicitly comparative, less exclusively human-centred, and more methodologically self-conscious about the concepts it employs.

**Acknowledgements** I am grateful to Dr Sebastian Sequoiah-Grayson, Prof. Flora Salim, and Sion Weatherhead (School of Computer Science and Engineering, UNSW Sydney) for helpful comments on an earlier draft; any errors remain my own.

**Author contributions** It is a sole-author article.

**Funding** Open Access funding enabled and organized by CAUL and its Member Institutions

**Data availability** No datasets were generated or analysed during the current study.

## Declarations

**Conflict of interest** The authors declare no conflict of interest.